\documentclass[10pt,twocolumn,letterpaper]{article}
\usepackage{iccv}
\newcommand{\nn}{AuxSegNet}

\usepackage{scalerel}
\usepackage{times}
\usepackage{epsfig}
\usepackage{graphicx}
\usepackage{amsmath}
\usepackage{amssymb}
\usepackage{float}
\usepackage{array}
\usepackage{adjustbox}
\usepackage{algorithmic}
\usepackage{algorithm}
\usepackage{multicol}
\usepackage{multirow}
\usepackage{booktabs}
\usepackage{array}
\usepackage{tabularx}
\usepackage{xcolor}
\usepackage{footmisc}
\usepackage{authblk}
\usepackage[colorlinks =true,
linkcolor = black,
urlcolor  = black,
citecolor = black,
anchorcolor = black]{hyperref}

\newcolumntype{L}[1]{>{\raggedright\let\newline\\\arraybackslash\hspace{0pt}}m{#1}}
\newcolumntype{C}[1]{>{\centering\let\newline\\\arraybackslash}m{#1}}
\newcolumntype{R}[2]{%
    >{\adjustbox{angle=#1,lap=\width-(#2)}\bgroup}%
    l%
    <{\egroup}%
}

\newcolumntype{B}{>{\arraybackslash\hsize=1.5\hsize}X}
\newcolumntype{S}{>{\centering\arraybackslash\hsize=0.2\hsize}X}
\newcolumntype{M}{>{\centering\arraybackslash}X}

\iccvfinalcopy 

\def\iccvPaperID{3388} 

\ificcvfinal\fi

\begin{document}

\title{Leveraging Auxiliary Tasks with Affinity Learning for Weakly Supervised Semantic Segmentation}
\author{
      Lian Xu$^1$,
      Wanli Ouyang$^2$,
      Mohammed Bennamoun$^1$,
      Farid Boussaid$^1$,
      Ferdous Sohel$^3$,
      and
      Dan Xu$^4$%
    \\\vspace{-6pt}
    $^1$The University of Western Australia \quad $^2$The University of Sydney \\
    $^3$Murdoch University \quad $^4$Hong Kong University of Sciences and Technology%
    \\
    \small{
    \tt{\{lian.xu,mohammed.bennamoun,farid.boussaid\}@uwa.edu.au}, \url{wanli.ouyang@sydney.edu.au},\\
    \url{F.Sohel@murdoch.edu.au}, \quad \url{danxu@cse.ust.hk}
    }
}

\maketitle
\ificcvfinal\thispagestyle{empty}\fi

\begin{abstract}
Semantic segmentation is a challenging task in the absence of densely labelled data. 
Only relying on class activation maps (CAM) with image-level labels provides deficient segmentation supervision.~Prior works thus consider pre-trained models to produce coarse saliency maps to guide the generation of pseudo segmentation labels. However, the commonly used off-line heuristic generation process cannot fully exploit the benefits of these coarse saliency maps. 
Motivated by the significant inter-task correlation, we propose a novel weakly supervised multi-task framework termed as \textbf{\nn}, to leverage saliency detection and multi-label image classification as auxiliary tasks to improve the primary task of semantic segmentation using only image-level ground-truth labels. 
Inspired by their similar structured semantics, we also propose to learn a cross-task global pixel-level affinity map from the saliency and segmentation representations.
The learned cross-task affinity can be used to refine saliency predictions and propagate CAM maps to provide improved pseudo labels for both tasks.
The mutual boost between pseudo label updating and cross-task affinity learning enables iterative improvements on segmentation performance.
Extensive experiments demonstrate the effectiveness of the proposed auxiliary learning network structure and the cross-task affinity learning method.
The proposed approach achieves state-of-the-art weakly supervised segmentation performance on the challenging PASCAL VOC 2012 and MS COCO benchmarks.~\footnote{https://github.com/xulianuwa/AuxSegNet}
\end{abstract}

\begin{figure}[t]
\setlength{\abovecaptionskip}{0cm}
\setlength{\belowcaptionskip}{0cm}
\begin{center}
\includegraphics[width=0.48\textwidth]{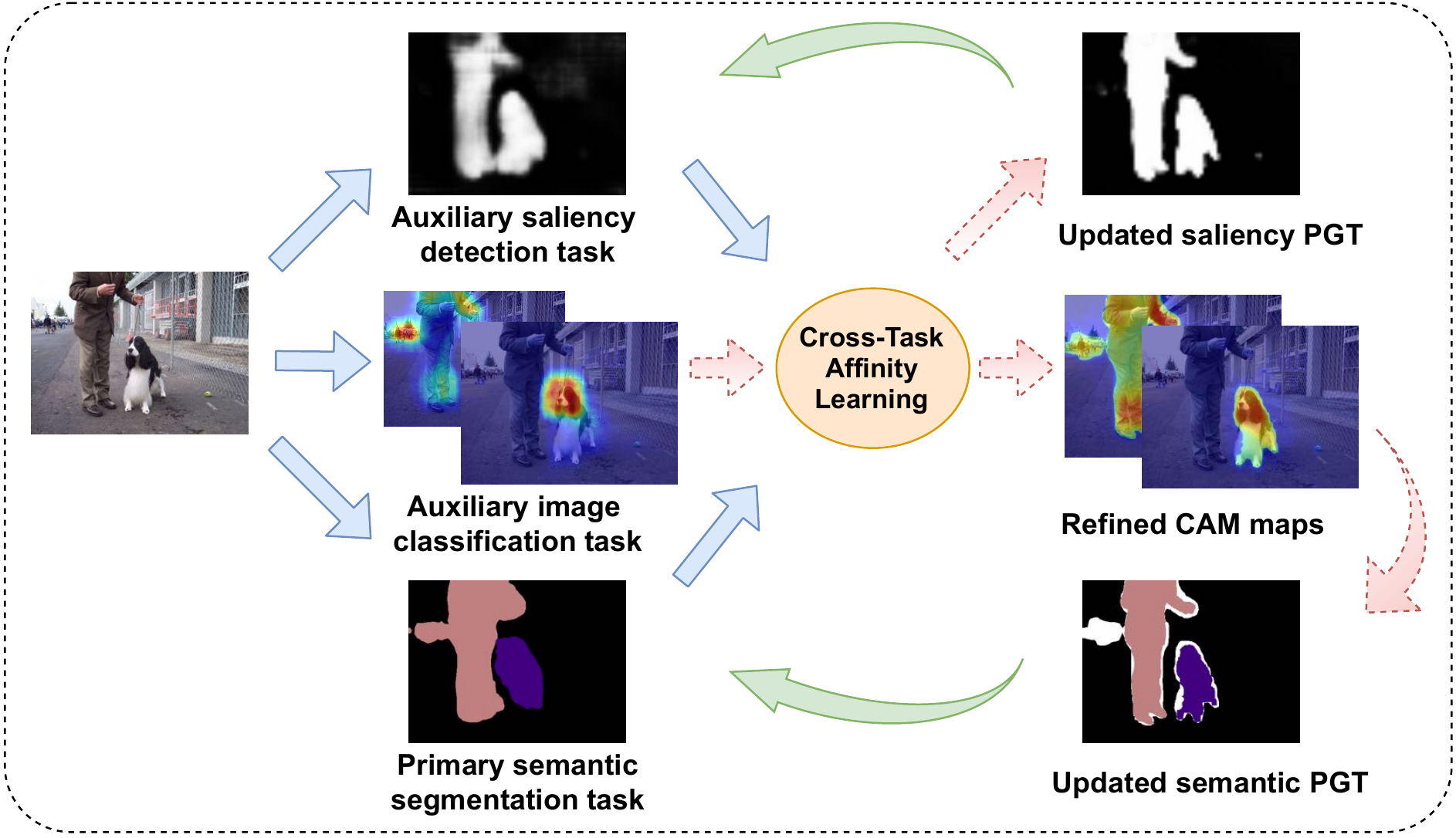}
\end{center}
  \caption{An illustration of the proposed approach for weakly supervised semantic segmentation. Our approach jointly learns two auxiliary tasks (\ie,~multi-label image classification and saliency detection) and a primary task (\ie,~semantic segmentation) only using image-level ground-truth labels, and performs affinity learning across two dense prediction tasks (\ie,~saliency detection and semantic segmentation). The learned affinity is then used to generate updated pseudo ground-truth (PGT) providing supervision to learn saliency detection and semantic segmentation. } 
\label{teaser}
\vspace{-3ex}
\end{figure}
\section{Introduction}
Semantic segmentation plays a vital role in many applications such as scene understanding and autonomous driving. It describes the process of assigning a semantic label to each pixel of an image. Prior works have achieved great success in the case of fully supervised semantic segmentation using Convolutional Neural Networks (CNNs). However, this has come at a high pixel-wise annotation cost. 
There has been an emerging research trend in semantic segmentation using less expensive annotations, such as
bounding boxes \cite{hu2018learning,song2019box}, scribbles \cite{lin2016scribblesup,tang2018normalized}, points and image-level labels \cite{pathak2015constrained,kolesnikov2016seed}.
Among them, image-level labels only indicate 
the presence or absence of objects in an image, resulting in an inferior segmentation performance compared to their fully supervised counterparts.

Most existing weakly supervised semantic segmentation (WSSS) approaches follow a two-step pipeline, \ie, generating pseudo segmentation labels and training segmentation models. A key element in generating pseudo segmentation labels is the class activation map (CAM) \cite{zhou2016learning}, which is extracted from CNNs trained on image-level labels. Although CAM maps identify class-specific discriminative image regions, those regions are quite sparse with very coarse boundaries.  
In order to generate high-quality pseudo segmentation labels, many approaches~\cite{wei2017object,wei2018revisiting,jiangintegral,wang2020self} have been presented to improve CAM maps from various aspects. 
Besides, existing methods~\cite{chaudhry2017discovering, hou2018self, sun2020mining,zhang2020splitting} typically use off-the-shelf saliency maps, combined with CAM maps, to determine reliable object and background regions. A general pre-trained saliency model can provide coarse saliency maps, which contain useful object localization information, on a target dataset.
However, in prior works, such coarse off-the-shelf saliency maps are only used as fixed binary cues in an off-line pseudo label generation process via heuristic thresholding, 
they are neither directly involved in the network training nor updated, which largely restricts their use to further benefit the segmentation performance.

Motivated by the observation that semantic segmentation, saliency detection and image classification are highly correlated, we propose a weakly supervised multi-task deep network (see Figure \ref{teaser}), which leverages saliency detection and multi-label image classification as auxiliary tasks to help learn the primary task of semantic segmentation. Through the joint training of these three tasks, an online adaptation can be achieved from pre-trained saliency maps to our target dataset. In addition, the task of saliency detection impels the shared knowledge to emphasize the difference between foreground and background pixels, thus driving the object boundaries of the segmentation outputs to coincide with those of the saliency outputs. Similarly, the image classification task highlights the discriminative features which can lead to more accurate segmentation predictions.  

Furthermore, we notice that, similar to these two dense prediction tasks, \ie, semantic segmentation and saliency detection, CAM maps also represent pixel-level semantics albeit they are only partially activated. Therefore, we propose to learn global pixel-level pair-wise affinities from the features of segmentation and saliency tasks to guide the propagation of CAM activations.~More specifically, two task-specific affinity maps are first learned for the saliency and segmentation tasks, respectively. To capture the complementary information between the two affinity maps, they are then adaptively integrated based on the learned self-attentions to produce a cross-task affinity map. 
Moreover, as we expect to learn semantic-aware and boundary-aware affinities so as to better update pseudo labels, we impose constraints on learning the cross-task affinities from task-specific supervision and joint multi-objective optimization. The learned cross-task affinity map is further utilized to refine saliency predictions and CAM maps to provide improved pseudo labels for both saliency detection and semantic segmentation respectively, enabling a multi-stage cross-task iterative learning and label updating.

In summary, the main contribution is three-fold:
\begin{itemize}
    \item We propose an effective multi-task auxiliary deep learning framework (\ie,~\nn) for weakly supervised semantic segmentation.~The proposed  \nn~leverages multi-label image classification and saliency detection as auxiliary tasks to help learn the primary task (\ie, semantic segmentation) using only image-level ground-truth labels. 

    \item We propose a novel method to learn cross-task affinities to refine both task-specific representations and predictions for semantic segmentation and saliency detection.
    The learned global pixel-level affinities can also be used to simultaneously update semantic and saliency pseudo labels in a joint cross-task iterative learning framework, yielding continuous boosts of the semantic segmentation performance. 

    \item Our proposed method achieves state-of-the-art results on PASCAL VOC 2012 and MS COCO datasets for the task of weakly supervised semantic segmentation.
\end{itemize}

\begin{figure*}[t]
\setlength{\abovecaptionskip}{0cm}
\setlength{\belowcaptionskip}{0cm}
\begin{center}
\includegraphics[width=0.95\textwidth]{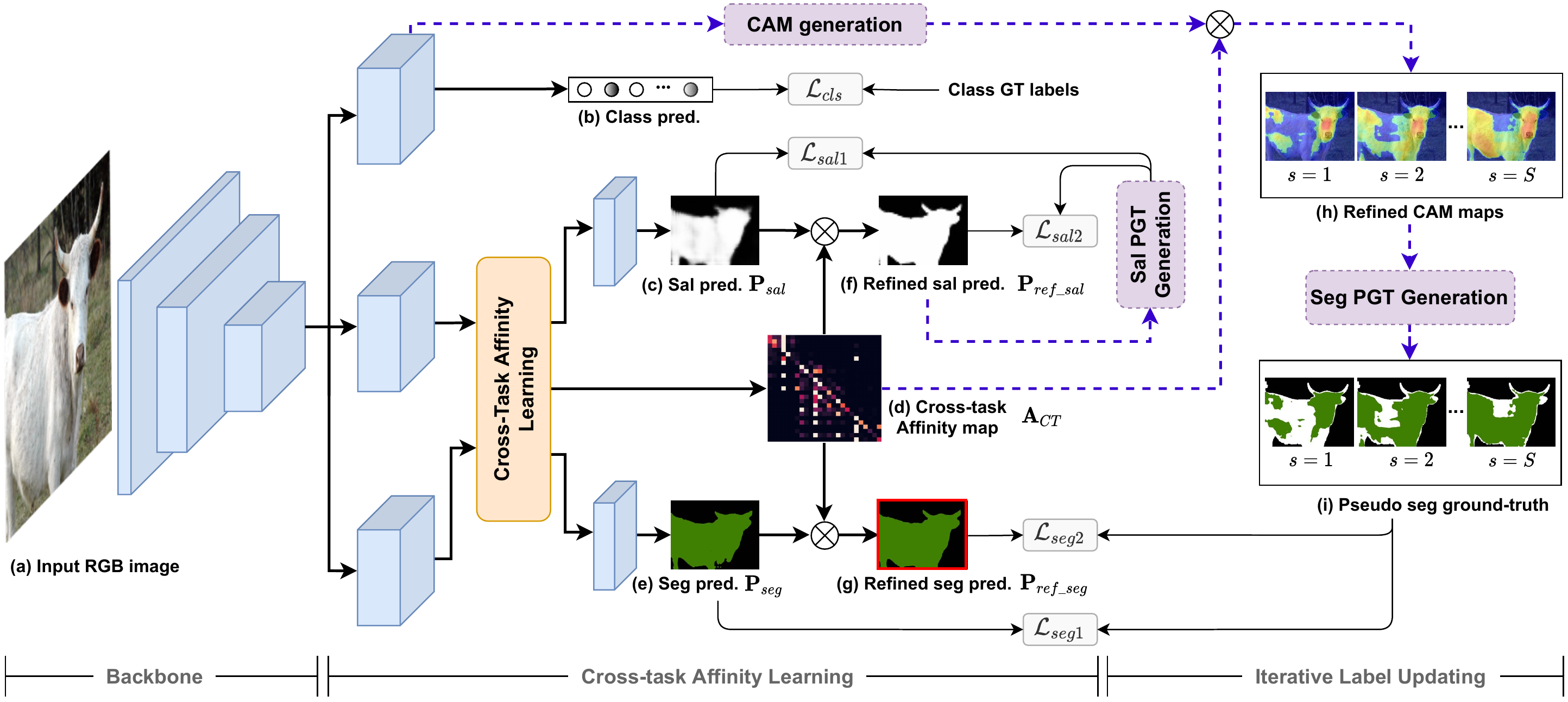}
\end{center}
 \vspace{-0.2cm}
  \caption{An overview of the proposed \nn. An input RGB image (a) is first passed through a backbone network to extract image features, which are then fed to three branches for multi-label image classification (b), saliency detection (c \& f) and semantic segmentation (e \& g), respectively. The proposed cross-task affinity learning module (see Figure \ref{nonlocal}) takes as inputs the segmentation and saliency feature maps, and outputs augmented feature maps for predicting both tasks (c \& e) and a cross-task affinity map (d) for task-specific prediction refinement (f \& g). The refined saliency predictions are used to update pseudo saliency labels, and the learned cross-task affinity map is used to refine CAM maps (h) to update pseudo segmentation labels (i) to retrain the network. The network training (black solid lines) and label updating (blue dashed lines) are performed alternatively for multiple stages (\ie,~$s=1,2,...,S$) to learn a more reliable affinity map and produce more accurate segmentation predictions. 
  } 
\label{overview}
\vspace{-1em}
\end{figure*}

\section{Related Work}
In this section, we review recent works from two closely related perspectives, \ie,~weakly supervised semantic segmentation and auxiliary learning for segmentation.

\par\noindent\textbf{Weakly supervised semantic segmentation.} 
Recent WSSS approaches commonly rely on CAM maps as seeds to generate pseudo segmentation labels. Several works focus on modifying classification objectives~\cite{wang2020self} or network architectures~\cite{wei2018revisiting} to improve CAM maps. A few works mine object regions based on erasing~\cite{wei2017object, hou2018self, li2018tell} and accumulation~\cite{jiangintegral} heuristics. Moreover, improved CAM maps can be achieved by exploring sub-category information~\cite{chang2020weakly} or mining cross-image semantics~\cite{fan2020cian, sun2020mining, li2020group}.

There are several methods \cite{wei2017object, wang2018weakly, wang2020weakly} which also perform iterative refinement on pseudo segmentation labels. Wei \etal~\cite{wei2017object} iteratively train the CAM network using images of which the discovered object parts are erased, to mine more object regions. This still results in combined CAM maps having coarse boundaries and non-discriminative object parts not activated. Wang \etal~\cite{wang2018weakly} improve pseudo segmentation labels by iteratively mining common object features from superpixels. Wang~\etal~\cite{wang2020weakly} use an affinity network, which learns local pixel affinities, to refine pseudo segmentation labels. The affinity network is iteratively optimized using the results from a segmentation network. In contrast to these methods requiring learning an additional single-modal affinity network to have alternating training with the segmentation network, we perform the cross-task affinity learning simultaneously with the proposed joint multi-task auxiliary learning network.

Several other works~\cite{ahn2018learning,fan2020cian} also refine pseudo labels by learning pixel affinities. However, Ahn~\etal~\cite{ahn2018learning} only learn pixel affinities based on the selected samples from the sparse CAM maps. Fan~\etal~\cite{fan2020cian} learn pixel affinity only to enhance feature representations for object estimation. In contrast, the proposed method is different from these approaches in the following aspects: it learns global pixel affinities across different tasks; it learns the pixel-level affinities to refine both task-specific representations and predictions; the affinity can be progressively improved along with more accurate saliency and segmentation results to be achieved with updated pseudo labels on both tasks.

\par\noindent\textbf{Auxiliary learning for segmentation}.
Many existing works have shown the effectiveness of multi-task learning~\cite{xu2018pad,Sheng2019Unsupervised}, which allows the sharing of learned knowledge across tasks to improve the performance of each individual task. 
In auxiliary learning, the goal of auxiliary tasks is to improve the performance of the primary task~\cite{XuMoving,liu2019self}.
For instance, Dai~\etal~\cite{dai2016instance} propose a multi-task network for instance segmentation by jointly learning to differentiate instances, estimate masks, and categorize objects. In \cite{chen2016dcan}, more accurate segmentation is achieved by learning an auxiliary contour detection task. In these cases, ground-truth labels are provided for both primary and auxiliary tasks.

In weakly supervised learning, the joint learning of object detection and semantic segmentation has been explored in~\cite{shen2019cyclic,kim2020weakly}. The joint learning of image classification and semantic segmentation has been investigated in~\cite{chaudhry2017discovering, zhang2019reliability, araslanov2020single}. Zeng~\etal~\cite{zeng2019joint} recently propose a joint learning framework for saliency detection and weakly supervised semantic segmentation, which however uses strong pixel-level saliency ground-truth labels as supervision.
In contrast, we leverage two auxiliary tasks (\textit{i.e.,} image classification and saliency detection) to facilitate the learning of the primary task of semantic segmentation using image-level ground-truth classification labels and off-the-shelf saliency maps. We take further advantage of multi-task features to learn cross-task affinities, which can improve the pseudo labels for both saliency and segmentation tasks simultaneously to achieve progressive boosts of the segmentation performance.

\section{The Proposed Approach}

\par\noindent\textbf{Overview}.~The overall architecture of the proposed \nn~is shown in Figure~\ref{overview}. An input RGB image is first fed into a shared backbone network. The generated backbone features are then forwarded to three task-specific branches which predict the class probabilities, a dense saliency map, and an dense semantic segmentation map, respectively.~The proposed cross-task affinity learning module first learns task-specific pixel-level pair-wise affinities, which are used to enhance the features of the saliency and segmentation tasks, respectively.
Then, the two task-specific affinity maps are adaptively integrated with the learned self-attentions as a guide to produce a global cross-task affinity map. This affinity map is further used to refine both the saliency and segmentation predictions during training. After each training stage, the learned cross-task affinity map is used to update the saliency and segmentation pseudo labels by refining the saliency predictions and the CAM maps. Only image-level ground-truth labels are required to train the proposed \nn.

\subsection{Multi-Task Auxiliary Learning Framework}
\par\noindent\textbf{Auxiliary supervised image classification}. The input image is first passed through the multi-task backbone network to produce a feature map $\mathbf{F} \in \mathbb{R}^{H\times W\times K}$, where $K$ is the number of channels, $H$ and $W$ are the height and width of the map, respectively. A Global Average Pooling (GAP) layer is then applied on $\mathbf{F}$ by aggregating each channel of $\mathbf{F}$ into a feature vector. 
After that, a fully connected (fc) layer is performed as a classifier to produce a probability distribution of the multi-class prediction. Given the weight matrix of the fc classifier $\mathbf{W} \in \mathbb{R}^{K\times C}$, with $C$ denoting the number of classes, the CAM map for a specific class $c$ at a spatial location $(i, j)$ can be formulated as $\mathbf{CAM}_c(i, j) = \sum_{k}^{K}\mathbf{W}_{k}^{c}\mathbf{F}_{k}(i,j)$
where $\mathbf{W}_{k}^{c}$ represents the weights corresponding to the class $c$ and the feature channel $k$, and $\mathbf{F}_{k}(i,j)$ represents the activation from the $k$-th channel of $\mathbf{F}$ at a spatial location $(i,j)$. The generated CAM maps are then normalized to be between 0 and 1 for each class $c$ by the maximum value in the two spatial dimensions.

\par\noindent\textbf{Auxiliary weakly supervised saliency detection with label updating.} For the saliency detection branch, feature maps from the backbone network are forwarded to two consecutive convolutional layers with dilated rates of 6 and 12, respectively.~The generated feature maps $\mathbf{F}_{sal\_in}$ are then fed to the proposed cross-task affinity learning module to obtain refined feature maps $\mathbf{F}_{sal\_out}$ and a global cross-task affinity map $\mathbf{A}_{CT}$. $\mathbf{F}_{sal\_in}, \mathbf{F}_{sal\_out} \in \mathbb{R}^{H\times W\times D}$ with $D$ denoting the number of channels. The refined feature maps are used to predict saliency maps $\mathbf{P}_{sal}$ by using a 1$\times$1 convolutional layer followed by a Sigmoid layer. The predicted saliency maps are further refined by the generated cross-task affinity map $\mathbf{A}_{CT}$ to obtain refined saliency predictions $\mathbf{P}_{ref\_sal}$. 
Since no saliency ground-truth is provided, we take advantage of pre-trained models which provide coarse saliency maps $\mathbf{Pt}_{sal}$ as initial pseudo labels. For the following training stages, we incorporate the refined saliency predictions of the previous stage (\ie,~stage $s-1$) to iteratively perform saliency label updates to continually improve the saliency learning as follows: 
\begin{equation}
      \mathbf{PGT}^{s}_{sal} = \begin{cases}
 \mathbf{\mathbf{Pt}}_{sal} & \text{ if } s = 0, \\ 
 \mathrm{CRF}_d(\frac{\mathbf{P}^{s-1}_{ref\_sal} + \mathbf{Pt}_{sal}}{2})& \text{ if } s > 0,
\end{cases}
\end{equation}
where $\mathbf{PGT}^{s}_{sal}$ denotes the saliency pseudo labels for the $\text{s}^{th}$ training stage, and $\mathrm{CRF}_d(\cdot)$ denotes a densely connected CRF following the formulation in~\cite{hou2019deeply} while using the average of $\mathbf{P}^{s-1}_{ref\_sal}$ and $\mathbf{Pt}_{sal}$ as a unary term.

\par\noindent\textbf{Primary weakly supervised semantic segmentation with label updating.} The semantic segmentation decoding branch shares the same backbone with the saliency decoding branch and the image classification branch. Similar to the saliency decoding branch, two consecutive atrous convolutional layers with rates of 6 and 12 are used to extract task-specific features $\mathbf{F}_{seg\_in}\in \mathbb{R}^{H\times W\times D}$, which are then fed to the cross-task affinity learning module. The output feature maps $\mathbf{F}_{seg\_out}$ are forwarded through a $1\times1$ convolutional layer and a Softmax layer to predict segmentation masks $\mathbf{P}_{seg}$, which are further refined by the learned cross-task affinity map to produce refined segmentation masks $\mathbf{P}_{ref\_seg}$.~To generate pseudo segmentation labels, we follow the conventional procedures \cite{wei2017object,wei2018revisiting,jiangintegral, hou2018self, lee2019ficklenet, sun2020mining} to select reliable object regions from CAM maps and background regions from off-the-shelf saliency maps \cite{hou2019deeply} by hard thresholding. More specifically, to generate the pseudo segmentation labels for the initial training stage (\ie,~stage 0), we first only train the classification branch using image-level labels to obtain the CAM maps. For the following training stages, we generate pseudo semantic labels by using the CAM maps refined by the cross-task affinities learned at the previous training stage.

\begin{figure}[t]
\setlength{\abovecaptionskip}{0cm}
\setlength{\belowcaptionskip}{0cm}
\begin{center}
\includegraphics[width=0.48\textwidth]{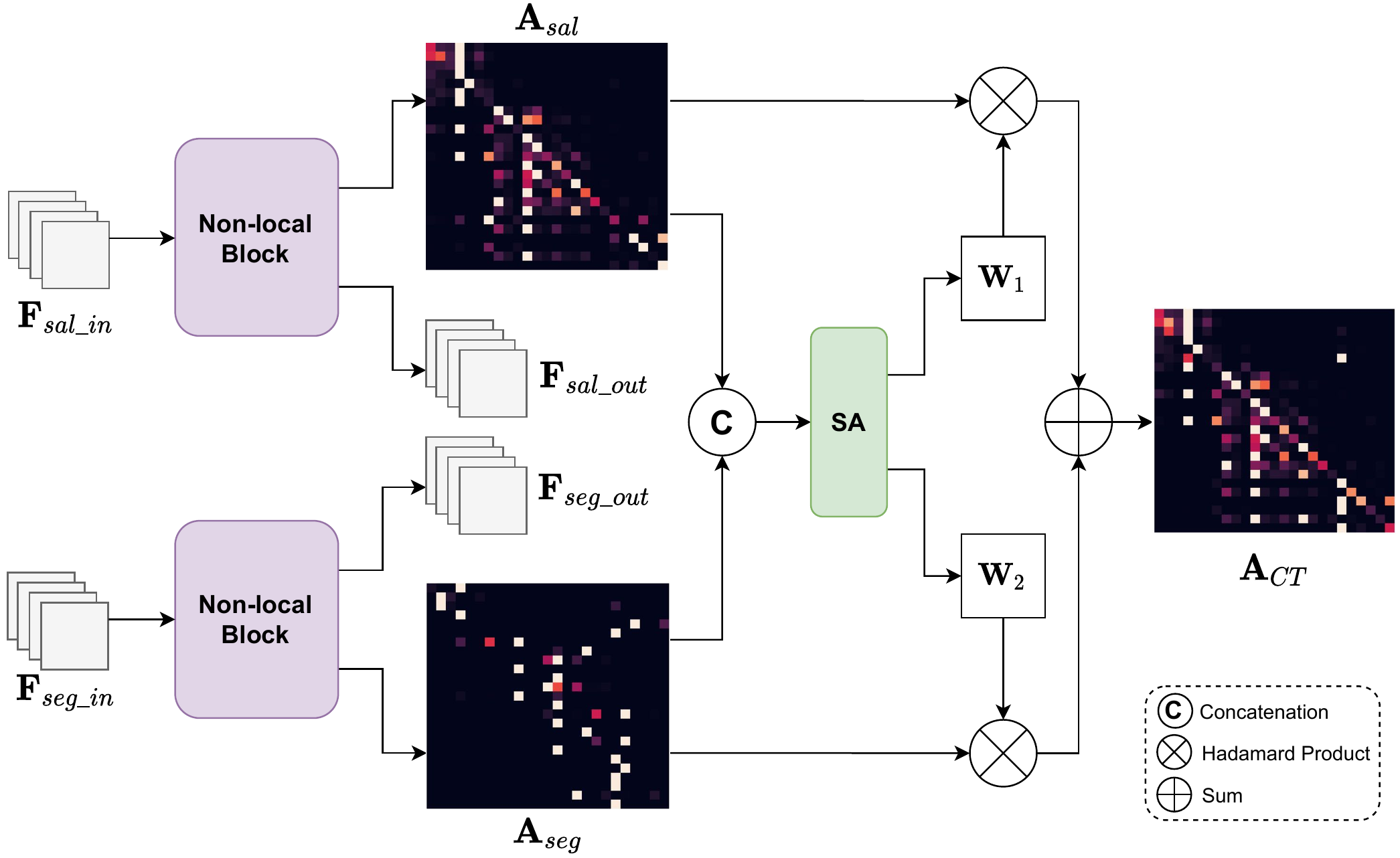}
\end{center}
 \vspace{-0.2cm}
  \caption{The detailed structure of the proposed cross-task affinity learning module. It consists of two non-local blocks which respectively learn task-specific saliency and segmentation affinity maps and refine their corresponding feature maps, and a self-attention (SA) module is designed to adaptively integrate two task-specific affinity maps to produce a global cross-task affinity map.} 
\label{nonlocal}
\vspace{-1em}
\end{figure}

\subsection{Cross-task Affinity Learning}

Given an input image, we can obtain task-specific features, \ie, $\mathbf{F}_{sal}$ and $\mathbf{F}_{seg}$ from the saliency and segmentation branches of the proposed \nn, respectively.
These feature maps for dense prediction tasks contain rich context information, and there are strong semantic relationships among the feature vectors in the spatial dimensions. Such beneficial information can be used to refine CAM maps as they share highly correlated semantic structures. 
\par\noindent\textbf{Cross-task affinity.} As illustrated in Figure~\ref{nonlocal}, we exploit non-local self-attention blocks (Non\-Local), which capture the semantic correlations of spatial positions based on the similarities between the feature vectors of any two positions, to learn task-specific pair-wise affinities for saliency detection and semantic segmentation tasks, respectively: 
\begin{align}
    [\mathbf{F}_{sal\_out},~\mathbf{A}_{sal}] &= \mathrm{Non\-Local}(\mathbf{F}_{sal\_in}), \\
    [\mathbf{F}_{seg\_out},~\mathbf{A}_{seg}] &= \mathrm{Non\-Local}(\mathbf{F}_{seg\_in}).
\end{align}
More specifically, given the saliency feature maps $\mathbf{F}_{sal\_in}\in \mathbb{R}^{H\times W\times D}$, we first use three $1\times1$ convolutional layers to transform it into a triplet of ($\mathbf{Q}$, $\mathbf{K}$, $\mathbf{V}$), which are then flattened in the spatial dimensions to be of size $HW\times D$. To compute the pair-wise affinity, we apply the dot product between each pair of entries of $\mathbf{Q}$ and $\mathbf{K}$ and obtain saliency affinity matrix $\mathbf{A}_{sal} \in \mathbb{R}^{HW\times HW}$, with each row representing the similarity values of a spatial position and the rest ones. We then apply the softmax operation along each row to normalize the similarity values to be between 0 and 1. Each position of the input feature maps is then modified by attending to all positions and taking the sum of the product of all positions with their corresponding affinity values associated to the given position in the feature space. The attended feature maps with aggregated context information are finally added to the input feature maps to form the output feature maps $\mathbf{F}_{sal\_out}$ with enhanced pixel-level representation. 
We apply the same non-local operation on the segmentation feature maps $\mathbf{F}_{seg\_in}$ to generate a segmentation affinity map $\mathbf{A}_{seg}$ and enhanced segmentation feature maps $\mathbf{F}_{seg\_out}$. 
To learn more consistent and accurate pixel affinities, we then integrate the two task-specific affinity maps by learning a self-attention (SA) module, 
which consists of two convolutional layers and a softmax layer.  
The generated two spatial attention maps from the SA module act as two weight maps, which are used to aggregate the segmentation and saliency affinity maps into one cross-task global affinity map $\mathbf{A}_{CT} \in \mathbb{R}^{HW\times HW}$ as follows:
\begin{align}
    [\mathbf{W_{1}}, \mathbf{W_{2}}] &= \mathrm{SA}(\mathrm{CONCAT}(\mathbf{A}_{sal}, \mathbf{A}_{seg})), \\
    \mathbf{A}_{CT} &= \mathbf{W}_{1}*\mathbf{A}_{sal} + \mathbf{W}_{2}*\mathbf{A}_{seg},
\end{align}
where  $\mathrm{CONCAT}$($\cdot$) denotes the concatenation operation, and $\mathbf{W}_{1}, \mathbf{W}_{2}\ \in \mathbb{R}^{HW\times HW}$ denote the learned two spatial self-attention maps as weight maps for the saliency and segmentation affinity maps, respectively. 
\par\noindent\textbf{Multi-task constraints on cross-task affinity.} To enhance the affinity learning, we consider imposing task-specific constraints on the generated cross-task affinity map. To this end, the generated cross-task affinity matrix is transposed and then enforced to refine both saliency and segmentation predictions during training as:
\begin{equation}
\setlength{\abovedisplayskip}{3pt}
\setlength{\belowdisplayskip}{3pt}
    \mathbf{P}_{ref\_sal}(i,j) = \sum_{k}^{H}\sum_{l}^{W}\mathbf{A}_{CT}(i,j,k,l)\cdot \mathbf{P}_{sal}(k,l),
\end{equation}
\begin{equation}
\setlength{\abovedisplayskip}{3pt}
\setlength{\belowdisplayskip}{3pt}
    \mathbf{P}_{ref\_seg}(i,j) = \sum_{k}^{H}\sum_{l}^{W}\mathbf{A}_{CT}(i,j,k,l)\cdot \mathbf{P}_{seg}(k,l).
\end{equation}
Then the learning of the cross-task affinity can gain effective supervision from both saliency and segmentation pseudo labels.~Therefore, the improvements on the updated pseudo labels can boost the affinity learning.

\subsection{Training and Inference}
\par\noindent\textbf{Overall optimization objective.} The overall learning objective function of \nn \ is the sum of the losses for the three tasks:
\begin{equation}
\begin{gathered}
    \mathcal{L}_{AuxSegNet} = \lambda_1\cdot\mathcal{L}_{cls} +   
    \lambda_2\cdot\mathcal{L}_{sal} + 
    \lambda_3\cdot\mathcal{L}_{seg}, \\
    \mathcal{L}_{sal}= \mathcal{L}_{sal1} + 
    \mathcal{L}_{sal2}, \\
     \mathcal{L}_{seg}= \mathcal{L}_{seg1} + 
    \mathcal{L}_{seg2}, 
\end{gathered}
\end{equation}
where $\mathcal{L}_{cls}$ is a multi-label soft margin loss computed between the predicted class probabilities and the image-level ground-truth labels to optimize the image classification network branch;
$\mathcal{L}_{sal1}$ and $\mathcal{L}_{sal2}$ are binary cross entropy losses computed between the predicted/refined saliency maps and the pseudo saliency label maps to optimize the saliency detection network branch and the affinity fusion module; $\mathcal{L}_{seg1}$ and $\mathcal{L}_{seg2}$ are pixel-wise cross entropy losses calculated between the predicted/refined segmentation masks and the pseudo segmentation label maps, and these losses optimize the segmentation branch and the affinity fusion module; and $\lambda1$, $\lambda2$ and $\lambda3$ are the loss weights for corresponding tasks.
\par\noindent\textbf{Stage-wise training.} We use a stage-wise training strategy for the entire multi-task network optimization. First, we only train the image classification branch with image-level labels for 15 epochs. The learned network parameters are then used as initialization to train the entire proposed \nn. We continually train the network for multiple stages with each training stage consisting of 10 epochs and update pseudo labels for saliency and segmentation branches after each training stage.
\begin{figure*}[t]
\setlength{\abovecaptionskip}{0cm}
\setlength{\belowcaptionskip}{0cm}
\begin{center}
\includegraphics[width=0.9\textwidth]{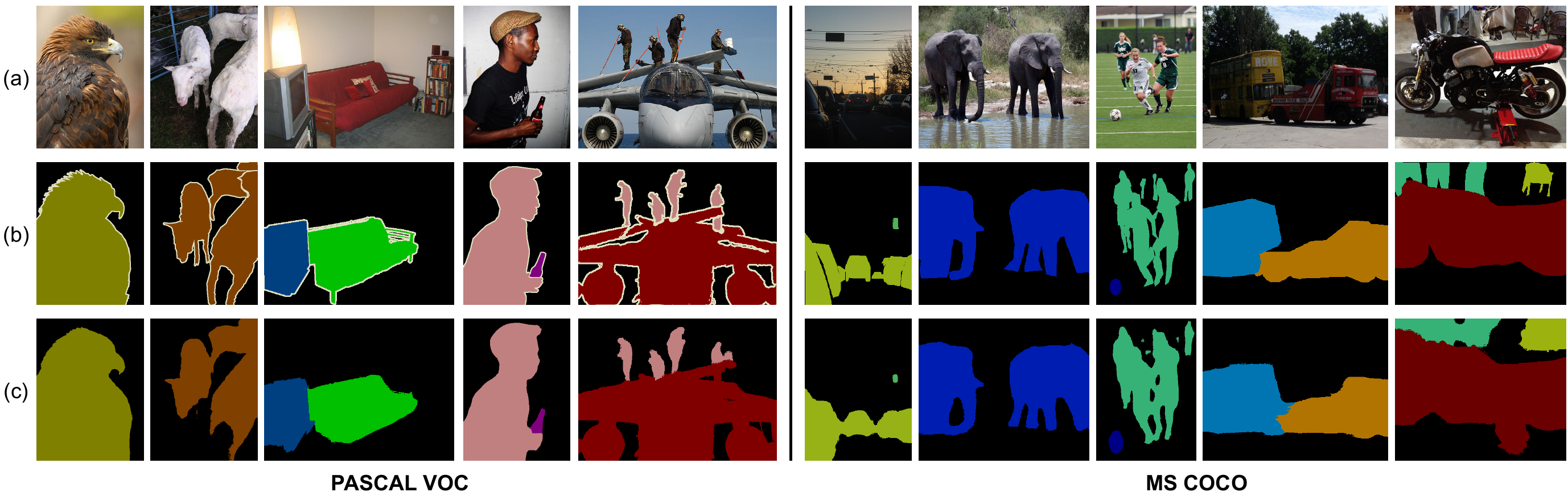}
\end{center}
 \vspace{-0.2cm}
  \caption{Qualitative segmentation results on the \textit{val} sets of PASCAL VOC and MS COCO. (a) Inputs. (b) Ground-truth. (c) Our results.} 
\label{segresults}
\vspace{-1em}
\end{figure*}
\par\noindent\textbf{Inference.} For inference, we use the segmentation prediction refined by the learned cross-task affinity map as the final segmentation results. 
\section{Experiments}

\subsection{Experimental Settings}
\par\noindent\textbf{Datasets.} To evaluate the proposed method, we conducted experiments on PASCAL VOC 2012 \cite{everingham2010pascal} and MS COCO datasets \cite{lin2014microsoft}. \textbf{PASCAL VOC} has 21 classes (including a background class) for semantic segmentation. This dataset has three subsets, \textit{i.e.,} training (\textit{train}), validation (\textit{val}) and test with 1,464, 1,449 and 1,456 images, respectively. Following common practice, \textit{e.g.}, \cite{chen2014semantic,kolesnikov2016seed}, we used additional data from \cite{hariharan2011semantic} to construct an augmented dataset with 10,582 images for training. \textbf{MS COCO} contains 81 classes (including a background class). It has 80K training images and 40K validation images. Note that only image-level ground-truth labels from these benchmarks are used in the training process. 
\begin{table}[t]
\caption{Performance comparison of WSSS methods in terms of mIoU(\%) on the PASCAL VOC 2012 \textit{val} and \textit{test} sets. $^{*}$: without post-processing. Sup.: supervision. I: image-level ground-truth. S: off-the-shelf saliency maps. $\text{S}^{'}$: saliency ground-truth.
\label{sota_res38}}
\centering
\small
\resizebox{1.0\linewidth}{!}{
\begin{tabular}{lcccc}
\toprule
Method            & Backbone  & Sup.          & Val            & Test           \\ \hline\hline
DSRG (CVPR18) \cite{huang2018weakly} &ResNet101 &I+S&61.4&63.2 \\
MCOF (CVPR18) \cite{wang2018weakly} & ResNet101 & I+S &60.3 & 61.2 \\
AffinityNet (CVPR18) \cite{ahn2018learning}&ResNet38&I&61.7&63.7 \\
SeeNet (NeurIPS18) \cite{hou2018self} &ResNet101&I+S&63.1 &62.8 \\
FickleNet (CVPR19)  \cite{lee2019ficklenet}  &      ResNet101           &  I+S&      64.9       & 65.3 \\
OAA$^{+}$ (ICCV19)  \cite{jiangintegral} &       ResNet101     & I+S&   65.2    &     66.4      \\
Zeng \etal (ICCV19) \cite{zeng2019joint}    & DenseNet                &I+$\text{S}^{'}$    &    63.3            &      64.3          \\ 
  CIAN (AAAI20)  \cite{fan2020cian}    &         ResNet101     &     I+S&  64.3         &  65.3 \\
  Zhang \etal (AAAI20) \cite{zhang2019reliability} & ResNet38 & I&62.6 & 62.9 \\
    Luo \etal (AAAI20)  \cite{luo2020learning}  &     ResNet101         &I     &       64.5       & 64.6 \\
Chang \etal (CVPR20) \cite{chang2020weakly}  &            ResNet101    &I     &      66.1        &    65.9\\ 
ICD (CVPR20) \cite{fan2020learning}    &          ResNet101          &   I+S&    67.8         &    68.0           \\ 
Araslanov \etal (CVPR20) \cite{araslanov2020single}    &          ResNet38    &I     &      62.7        &   64.3       \\ 
SEAM (CVPR20) \cite{wang2020self} & ResNet38 &I& 64.5 & 65.7 \\
 Zhang \etal (ECCV20) \cite{zhang2020splitting}   &       ResNet50  &   I+S&   66.6         &   66.7    \\  
 Sun \etal (ECCV20) \cite{sun2020mining}   &          ResNet101        &I+S&        66.2          &      66.9        \\
 CONTA (NeurIPS20) \cite{zhang2020causal}   &          ResNet38         &  I&    66.1       &    66.7            \\  
 \hline
   \textbf{\nn}$^{*}$ (Ours) & ResNet38 & I+S&   66.1 &  66.1 \\
  \textbf{\nn} (Ours) & ResNet38 &I+S&    \textbf{69.0} &  \textbf{68.6}\\
  \bottomrule
\end{tabular}
}
\vspace{-2ex}
\end{table}
\begin{table}[t]
\caption{Performance comparison of WSSS methods in terms of mIoU(\%) on the MS COCO \textit{val} set. }
\label{coco}
\small
\centering
\begin{tabular}{lccc}
\toprule 
Method            & Backbone  &Sup.          & Val                       \\ \hline\hline
SEC (CVPR16) \cite{kolesnikov2016seed} & VGG16 & I+S &22.4 \\
DSRG (CVPR18) \cite{huang2018weakly} & VGG16 & I+S&26.0 \\
Wang \etal (IJCV20) \cite{wang2020weakly}&VGG16&I&27.7\\
    Luo \etal (AAAI20)  \cite{luo2020learning}  &     VGG16              &  I&    29.9 \\
SEAM (CVPR20) \cite{wang2020self} & ResNet38 & I&31.9 \\
CONTA (NeurIPS20) \cite{zhang2020causal} & ResNet38 & I&32.8 \\
 \hline
  \textbf{\nn} (Ours) & ResNet38 &I+S & \textbf{33.9} \\
  \bottomrule 
\end{tabular}
\vspace{-4ex}
\end{table}

\par\noindent\textbf{Evaluation metrics.}
As in previous works \cite{jiangintegral,lee2019ficklenet,wei2018revisiting,huang2018weakly,ahn2018learning,wang2018weakly}, we used the mean Intersection-over-Union (mIoU) of all classes between the segmentation outputs and the ground-truth images to evaluate the performance of our method on the \textit{val} and \textit{test} sets of PASCAL VOC and the \textit{val} set of MS COCO. We also used precision, recall and mIoU to evaluate the quality of the pseudo segmentation labels. The results on the PASCAL VOC \textit{test} set were obtained from the official PASCAL VOC online evaluation server.
\par\noindent\textbf{Implementation details.}
In our experiments, models were implemented in PyTorch. 
We use ResNet38 \cite{wu2019wider,ahn2018learning} as the backbone network. 
For data augmentation, we used random horizontal flipping, random cropping to size $321\times 321$ and color jittering. The polynomial learning rate decay was chosen with an initial learning rate of 0.001 and a power of 0.9. We used the stochastic gradient descent (SGD) optimizer to train \nn~with a batch size of 4.
At inference, we use multi-scale testing and CRFs with hyper-parameters suggested in \cite{chen2014semantic} for post-processing.

\subsection{Comparison with State-of-the-arts}

\par\noindent\textbf{PASCAL VOC.} We compared the segmentation performance of the proposed method with state-of-the-art WSSS approaches. Table \ref{sota_res38} shows that the proposed method achieves mIoUs of 69.0\% and 68.6\% on the \textit{val} and \textit{test} sets, respectively. 
Our method outperforms the recent methods \cite{fan2020cian, fan2020learning, sun2020mining, zhang2020splitting} using off-the-shelf saliency maps by 4.7\%, 1.2\%, 2.8\% and 2.4\%, respectively, on the PASCAL VOC \textit{val} set.
The qualitative segmentation results on PASCAL VOC \textit{val} set are shown in Figure \ref{segresults} (left). Our segmentation results are shown to adapt well to different object scales and boundaries in various and challenging scenes.

\par\noindent\textbf{MS COCO.} To demonstrate the generalization ability of the proposed method, we also evaluated our method on the challenging MS COCO dataset. Table \ref{coco} shows segmentation results of recent methods on the \textit{val} set, where the result of SEAM \cite{wang2020self} is from the re-implementation by CONTA \cite{zhang2020causal}. Our method achieves an mIoU of 33.9\%, which is superior to state-of-the-art methods. Figure \ref{segresults} (right) presents several examples of qualitative segmentation results, 
which indicate that our proposed method performs well in different complex scenes, such as small objects or multiple instances.

\subsection{Ablative Analysis}
\begin{table}[t]
\caption{Performance comparison of jointly learning different combinations of multiple tasks in terms of mIoU(\%) on PASCAL VOC 2012 \textit{val} set. $Seg., Cls.$, and $Sal.$ denote semantic segmentation, image classification, and saliency detection, respectively.}
\label{mt_results}
\centering
\small
\begin{tabular}{lcccc}
\toprule 
\multirow{2}{*}{Config}&\multicolumn{3}{c}{Branches}&\multirow{2}{*}{mIoU} \\\cline{2-4}
&Seg.&Cls.&Sal.&\\
\hline\hline
Baseline&\checkmark&&&56.9 \\
\nn~(w/ seg., cls.)&\checkmark&\checkmark&&57.6 \\
\nn~(w/ seg., sal.)&\checkmark&&\checkmark&59.8\\
\nn~(w/ seg., cls., sal.)&\checkmark&\checkmark&\checkmark&\textbf{60.8}\\

\bottomrule
\end{tabular}
\end{table}

\par\noindent\textbf{Effect of auxiliary tasks.} We compared results from the one-branch baseline model for semantic segmentation to several different variants: (\textbf{i}) baseline + cls: leveraging multi-label image classification, (\textbf{ii}) baseline + sal: leveraging saliency detection, and (\textbf{iii}) baseline + cls + sal: leveraging both image classification and saliency detection. Several conclusions can be drawn from Table \ref{mt_results}. Firstly, the baseline performance with the single task of semantic segmentation is only 56.9\%. Joint learning an auxiliary task of either image classification or saliency detection both improve the segmentation performance significantly. In particular, learning saliency detection brings a larger performance gain of around 3\%. Furthermore, leveraging both auxiliary tasks yields the best mIoU of 60.8\% without using any post-processing. This indicates that these two related auxiliary tasks can improve the representational ability of the network to achieve more accurate segmentation predictions in the weakly supervised scenario. 

\begin{table}[t]
\caption{Comparison of affinity learning with different settings in terms of mIoU(\%) on PASCAL VOC 2012 \textit{val} set. CT: cross-task.}
\label{aff_results}
\centering
\small
\begin{tabular}{lc}
\toprule 
{Config}&mIoU \\
\hline\hline
\nn~(multi-task baseline)&60.8 \\
+ Seg. affinity with seg. constraint&61.5\\
+ CT affinity with seg. constraint&62.6\\
+ CT affinity with seg. and sal. constraints &\textbf{64.1}\\
\bottomrule
\end{tabular}
\end{table}

\begin{table}[t]
\caption{Segmentation performance of the proposed \nn~after different training stages in terms of mIoU (\%) on PASCAL VOC 2012 \textit{val} set. Stage 0 denotes the training stage with the initial pseudo segmentation ground-truth without refinement.}
\label{iterative_results}
\centering
\small
\begin{tabular}{lccccc}
\toprule 
&Stage0&Stage1&Stage2&Stage3&Stage4\\
\hline\hline
mIoU&64.1&65.6&66.0&\textbf{66.1}&\textbf{66.1} \\
\bottomrule
\end{tabular}
\vspace{-2em}
\end{table}

\begin{figure}[t]
\setlength{\abovecaptionskip}{0cm}
\setlength{\belowcaptionskip}{0cm}
\begin{center}
\includegraphics[width=0.4\textwidth]{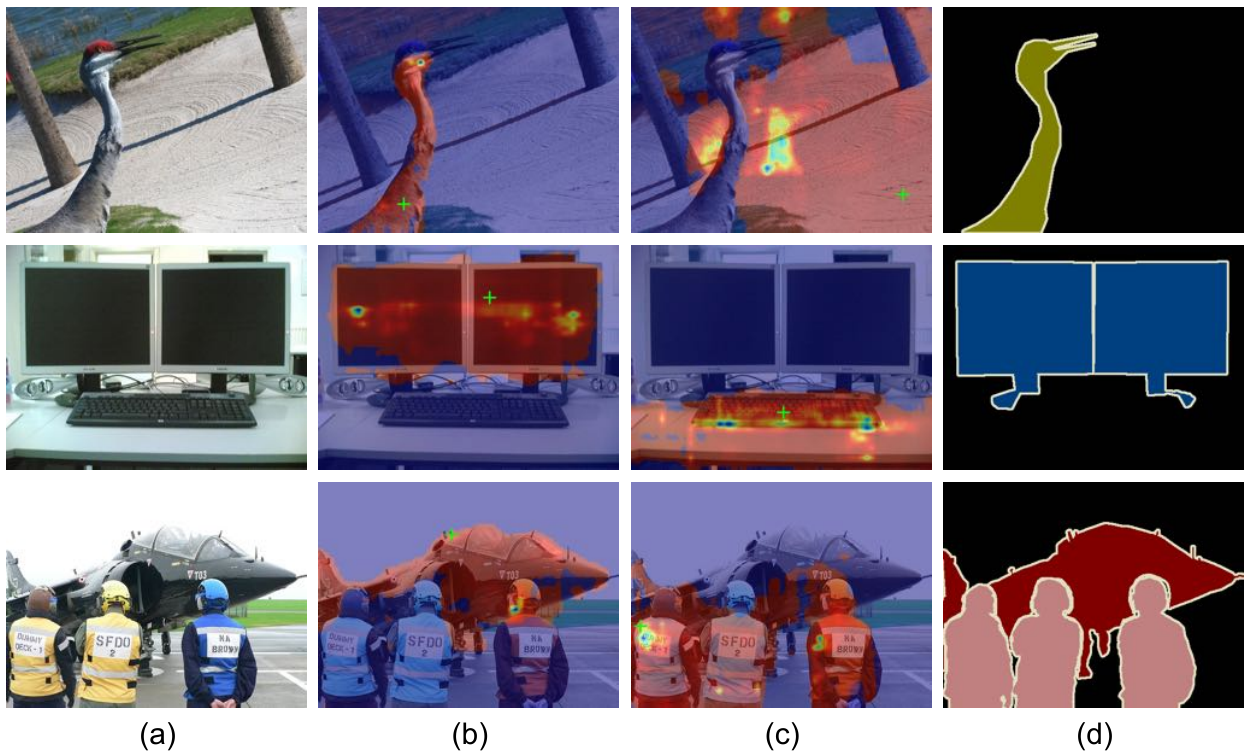}
\end{center}
 \vspace{-0.2cm}
  \caption{Visualization of the learned cross-task affinity maps of two selected points in the images on PASCAL VOC \textit{train} set. (a) Inputs. (b)-(c) Two learned cross-task affinity maps for two points marked by the green crosses. (d) Segmentation ground-truth.} 
\label{affmap}
\vspace{-1em}
\end{figure}

\par\noindent\textbf{Different settings for affinity learning.} Table \ref{aff_results} shows ablation studies on the impact of different affinity learning settings on the segmentation performance. Without affinity learning, the segmentation mIoU is only 60.8\%. The performance is improved to 61.5\% by only learning segmentation affinity to refine segmentation predictions. Learning a cross-task affinity map which integrates both segmentation and saliency affinities brings a further performance boost of 1.1\%. By forcing the cross-task affinity map to learn to refine both segmentation and saliency predictions, our model attains a significant improvement, reaching an mIoU of 64.1\%. This demonstrates the positive effect of the multi-task constraints on learning pixel affinities to enhance weakly supervised segmentation performance.

\begin{figure}[t]
\setlength{\abovecaptionskip}{0cm}
\setlength{\belowcaptionskip}{0cm}
\begin{center}
\includegraphics[width=0.35\textwidth]{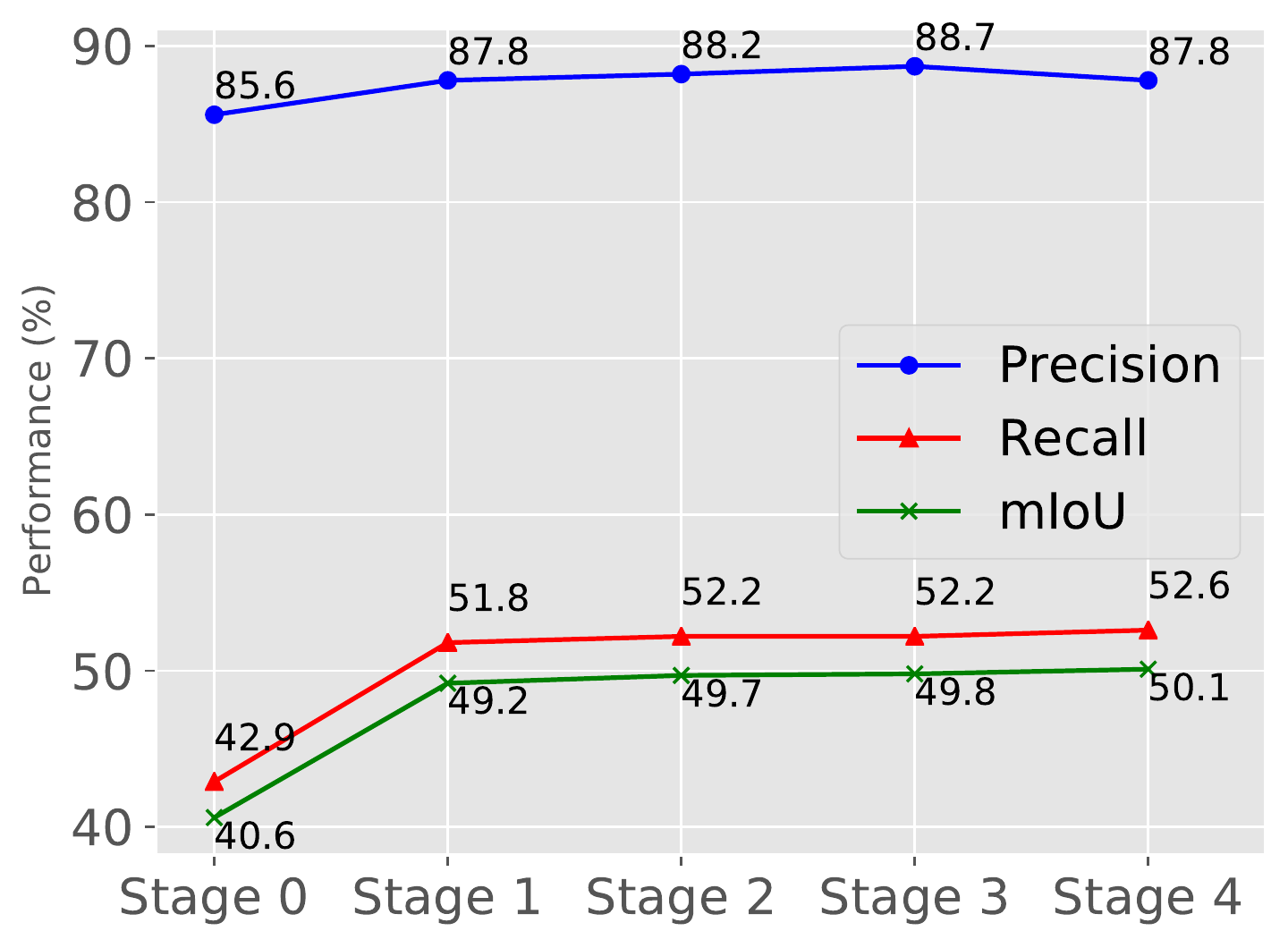}
\end{center}
 \vspace{-0.2cm}
  \caption{Evaluation of pseudo segmentation labels (PGT) for each training stage in terms of precision(\%), recall(\%) and mIoU(\%) on PASCAL VOC 2012 \textit{train} set.} 
\label{pgt_line}
\vspace{-1em}
\end{figure}

\begin{figure*}[t]
\begin{center}
\includegraphics[width=0.9\textwidth]{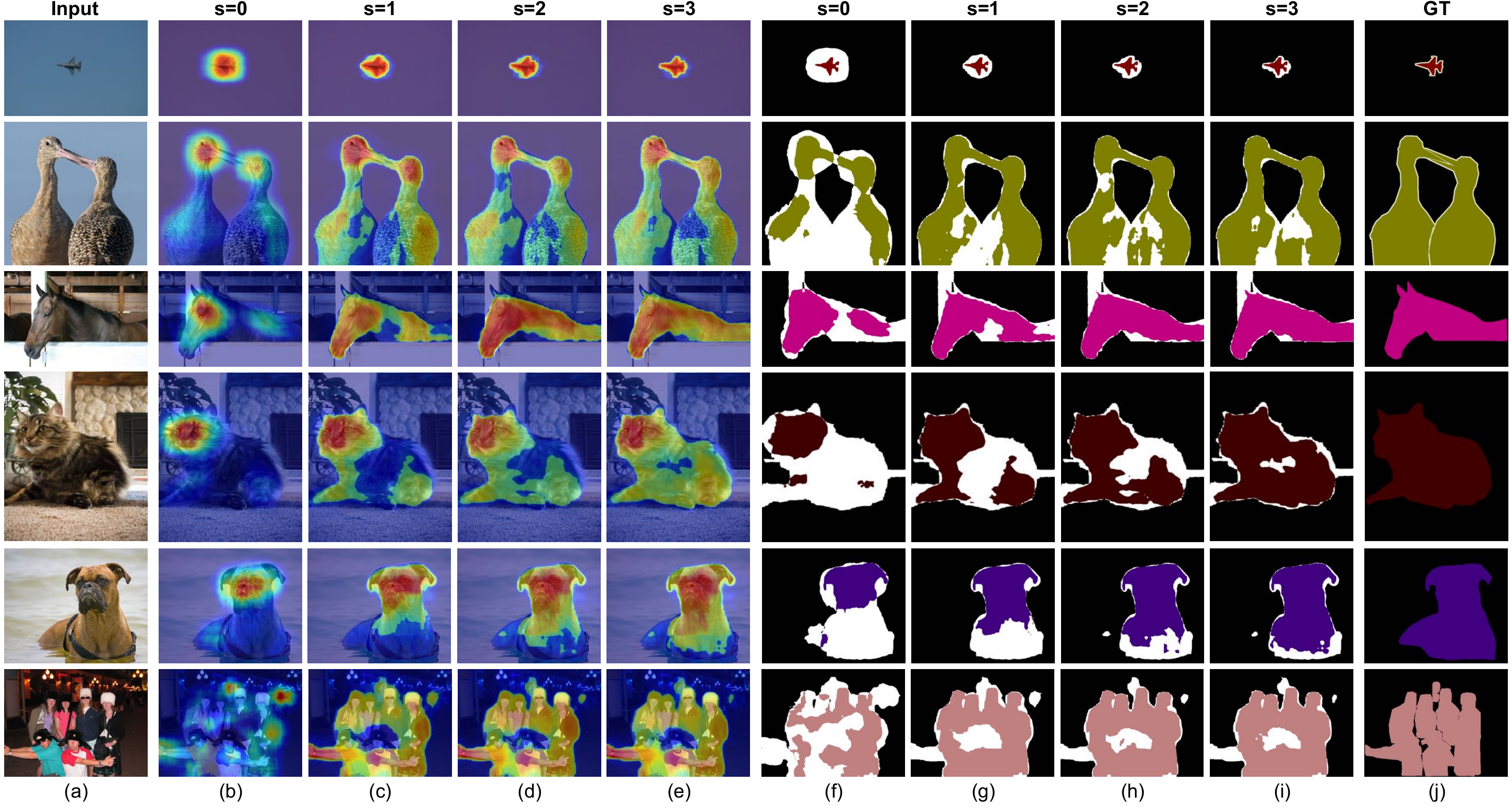}
\end{center}
\vspace{-0.3cm}
  \caption{Visualization of CAM maps and pseudo segmentation labels with iterative improvements. (a) Inputs. (b) Initial CAM maps without refinement. (c)-(e) Refined CAM maps used to generate pseudo segmentation labels for Stage 1 to Stage 3. (f) Initial pseudo semantic labels for Stage 0. (g)-(i) Pseudo segmentation labels for Stage 1 to Stage 3. (j) Segmentation ground-truth. With the iterative cross-task affinity learning, the refined CAM maps become more complete with more accurate boundaries and the generated corresponding pseudo segmentation labels are closer to the ground-truth in terms of precision and recall.} 
\label{cam_pgt}
\vspace{-0.4cm}
\end{figure*}

Figure \ref{affmap} presents several examples of the learned cross-task affinity maps for two selected points in each image. The affinity map for each pixel in an image is of the image size and it represents the similarity of this pixel to all pixels in the image. We can observe that, in the first row, the affinity map of the point labelled as ``bird" highlights almost the entire object region although this point is far from the most discriminative object part ``head".
Moreover, in the second row, the learned affinity map for the point belonging to the monitor activates most regions of the two monitor instances while it does not respond to the ``keyboard" region which is similar in color in the background, and vice versa. In the third row, the affinity maps for the ``airplane" and "person" points both present good boundaries.

\par\noindent\textbf{Iterative improvements.} To validate the effectiveness of the proposed iterative cross-task affinity learning, we evaluated the quality of the generated pseudo segmentation labels for each training stage. As shown in Figure \ref{pgt_line}, the precision, recall and mIoU of the initial pseudo segmentation labels generated by the CAM maps without refinement are 85.6\%, 42.9\% and 40.6\%, respectively. After the first round of affinity learning, the updated pseudo labels for Stage 1 are significantly improved by 2.2\%, 8.9\% and 8.6\% on these three metrics by using the refined CAM maps. Another round of affinity learning further boosts pseudo labels to 88.2\%, 52.2\% and 49.7\% on the three metrics. In the following training stages, pseudo segmentation labels are slightly improved, and they are saturated at Stage 3 as we can observed that the precision starts to drop at Stage 4. As shown in Table \ref{iterative_results}, the segmentation performance presents consistent improvements as pseudo labels update after each training stage. Overall, the proposed \nn~attains a reasonable improvement of 2\% by using iterative label updates with the learned cross-task affinity, reaching the best mIoU of 66.1\% without any post-processing.

To qualitatively evaluate the benefits of the proposed iterative affinity learning, Figure \ref{cam_pgt} presents several examples of CAM maps and the corresponding generated pseudo segmentation labels, and their iterative improvements along the training stages. As shown in Figure \ref{cam_pgt} (b), the CAM maps without affinity refinement for the initial training stage are either over-activated for small-scale objects, sparse for multiple instances or they only focus on the local discriminative object parts for large-scale objects. By refining CAM maps with the cross-task affinity learned at Stage 0, Figure \ref{cam_pgt} (c) shows that more object regions are activated for large-scale objects and the CAM maps for small-scale objects are more aligned to object boundaries. With more training stages, as shown in Figure \ref{cam_pgt} (d)-(e), the refined CAM maps become more integral with more accurate boundaries, which is attributed to the more reliable affinity learned with iteratively updated pseudo labels. The generated pseudo segmentation labels are shown to become progressively improved in Figure \ref{cam_pgt} (f)-(i), and they are close to the ground-truth labels. 

\begin{table}[t]
\caption{Segmentation results in terms of mIoU (\%) using saliency models pre-trained in supervised or weakly supervised forms on PASCAL VOC 2012 \textit{val} set. $^{*}$ denotes `without post-processing'.}
\resizebox{1.0\linewidth}{!}{
\begin{tabular}{lccccc}
\toprule
Pre-trained saliency models & Baseline & Final$^*$ & $\Delta$ & Final \\ \hline\hline
Zhang~\textit{et al.} (CVPR20)~\cite{zhang2020weakly}        &       53.0   &   63.7    &   10.7    & 66.5      \\
PoolNet (CVPR19)~\cite{liu2019simple}   &          55.7        &     65.7  &10.0 &68.4   \\
MINet (CVPR20)~\cite{pang2020multi}   &             55.8     &     66.6&10.8 & 68.9\\
Ours with DSS~\cite{hou2019deeply}          &       56.9   &  66.1     &  9.2     &     69.0 \\
\bottomrule
\end{tabular}
}
\label{fullsal}
\vspace{-2em}
\end{table}

\par\noindent\textbf{Different pre-trained saliency models.} To evaluate the sensitivity to the pre-trained saliency models, we conducted experiments using different pre-trained saliency models, \textit{i.e.,} Zhang~\etal~\cite{zhang2020weakly}, PoolNet~\cite{liu2019simple} used in~\cite{sun2020mining,zhang2020splitting}, and MINet~\cite{pang2020multi}, in which the first and the other two are in the weakly supervised and fully supervised forms, respectively. As shown in Table~\ref{fullsal}, our method achieves comparable results with these different saliency inputs, verifying the generalization ability of our method to different pre-trained saliency models. Moreover, with all these different pre-trained saliency models, our method can consistently produce significant performance improvements (see $\Delta$) over the baseline, which further confirmed the effectiveness of the proposed method.
\section{Conclusion}
In this work, we proposed to leverage auxiliary tasks without requiring additional ground-truth annotations for WSSS. More specifically, we proposed \nn~with a shared backbone and two auxiliary modules performing multi-label image classification and saliency detection to regularize the feature learning for the primary task of semantic segmentation. 
We also proposed to learn a cross-task pixel affinity map 
from saliency and segmentation feature maps. The learned cross-task affinity can be used to refine saliency predictions and CAM maps to provide improved pseudo labels for both tasks, which can further guide the network to learn more reliable pixel affinities and produce more accurate segmentation predictions. Iterative training procedures were thus conducted and realized progressive improvements on the segmentation performance.
Extensive experiments on the challenging PASCAL VOC 2012 and MS COCO demonstrate the effectiveness of the proposed method and establish new state-of-the-art results.
\vspace{0.1pt}
\par\noindent\textbf{Acknowledgment.} This research is supported in part by Australian Research Council Grant DP150100294, DP150104251, DP200103223, Australian Medical Research Future Fund MRFAI000085, the Early Career Scheme of the Research Grants Council (RGC) of the Hong Kong SAR under grant No. 26202321 and HKUST Startup Fund No. R9253.

{\small
\bibliographystyle{ieee_fullname}
\bibliography{egbib}
}

\end{document}